\documentclass[runningheads]{llncs}
\usepackage[usenames,dvipsnames]{color}

\usepackage{amsmath}
\usepackage{amssymb}
\usepackage{bm}
\usepackage{graphicx}
\usepackage{svg}
\usepackage{float}
\usepackage{booktabs}
\usepackage{multirow}
\usepackage[hyphens]{url}
\usepackage[hidelinks,breaklinks]{hyperref}

\newcommand*{\placeholder}{\makebox[1ex]{$\bm{\cdot}$}}

\begin{document}
\title{Learning Low-Level Causal Relations using a Simulated Robotic Arm}
\author{
Miroslav Cibula\inst{1} \and
Matthias Kerzel\inst{2} \and
Igor Farkaš\inst{1}
}
\authorrunning{M.~Cibula et al.}
\institute{
Faculty of Mathematics, Physics and Informatics, Comenius University Bratislava, Bratislava, Slovakia \\
\email{cibula25@uniba.sk, igor.farkas@fmph.uniba.sk}
\and Department of Informatics, University of Hamburg, Hamburg, Germany \\
\email{matthias.kerzel@uni-hamburg.de}
}
\maketitle              %
\begin{abstract}
Causal learning allows humans to predict the effect of their actions on the known environment and use this knowledge to plan the execution of more complex actions. Such knowledge also captures the behaviour of the environment and can be used for its analysis and the reasoning behind the behaviour. This type of knowledge is also crucial in the design of intelligent robotic systems with common sense.
In this paper, we study causal relations by learning the forward and inverse models based on data generated by a simulated robotic arm involved in two sensorimotor tasks. As a next step, we investigate feature attribution methods for the analysis of the forward model, which reveals the low-level causal effects corresponding to individual features of the state vector related to both the arm joints and the environment features. This type of analysis provides solid ground for dimensionality reduction of the state representations, as well as for the aggregation of knowledge towards the explainability of causal effects at higher levels. 

\end{abstract}

\section{Introduction}

Observing and learning causal relations in a given environment is an essential element of cognition in humans and other high animals. Thanks to this ability, agents can form intuitive theories from multiple observations and use them to predict the environment's behaviour in response to their actions \cite{Gerstenberg2017}. 
This common sense understanding includes the knowledge of intuitive physics, a key ingredient of early cognitive development \cite{Lake2017}.

Causal models capturing and learning causal relationships from observations can be used for action planning towards task completion. Analysis of these models, encapsulating intuition about the given environment and their predictions, can be helpful for causal reasoning. In humans, a seven-grade model of the evolution of causal cognition has been proposed, with increasingly more complex causal skills \cite{Lombard2017}. These range from individual causal understanding and tracking behaviour (understanding perceived effects of one's own motor actions) up to what the authors call causal network understanding (e.g. after learning that wind can cause an apple to fall, the person may understand that wind can also cause other things to fall or move) \cite{Gardenfors2018}.

Studying causation is a complex area, and it is being transferred to robotics, which essentially involves embodied agents (robots) interacting with the world \cite{Hellstroem2021}. The authors proposed an analysis of the role of causal reasoning in robotics, organized into two parts. The first part is a novel categorization of robot causal cognition inspired by the categorization of human causal cognition \cite{Lombard2017}. The latter describes a hierarchy of seven grades of causal skills, with humans mastering all grades and animals only certain grades, all according to their stage in evolution. 
They define eight categories of robot causal cognition at the sense–plan–act level, divided into three groups: learning causal relations, inferring the causes related to an interacting human, and a robot deciding how to act.
In our work, we focus on two categories of learning causal relations, as explained below.

Studying causality is an emerging area of machine learning research \cite{Zhang2018,Schoelkopf2022}, fueled by inspiration from the pioneering work on graphical causal inference by Judea Pearl \cite{Pearl2009} and collaborators \cite{Peters2017}. It has been argued that understanding causality can be beneficial for machine learning and artificial intelligence, leveraging it toward building more robust models with common sense \cite{Schoelkopf2022,Zhu2020}.
In machine learning, the conceptual framework deals with formal systems that can be studied at four levels, as proposed in \cite{Schoelkopf2022}, ranging from the most detailed mechanistic/physical level, through structural causal, causal graphical, up to (most abstract) statistical level.

The ideas of causal mechanisms are also naturally usable in robotics, where these levels apply. In our work, we focus on the lowest, mechanistic level corresponding to individual components of the state space representation, including the robot's individual joints and features of the world.
Our work offers two main contributions: (1) we explore causal learning using forward and inverse models encapsulating intuitive knowledge about the environment and train them on synthetic data generated by motor babbling in simulation, and (2) we analyze these models to extract information about the behaviour of the environment.

\section{Background and Related Work}

The sensorimotor knowledge in a robotic system is commonly represented by a pair (or pairs of) of complementary models: the {\em forward model} (FM) that predicts sensory consequences of one's own actions, and an {\em inverse model} (IM) that predicts actions in order to reach the desired state \cite{Wolpert1998}.\footnote{More recent approaches based on predictive coding reconceptualize the cognitive view on sensorimotor behaviour and question the need for having an IM at all, arguing that instead, a single, integrative forward model is sufficient \cite{Ciria21}. In our work, we adhere to the standard, two-models view.} 

The FM is commonly called a causal model, which is mathematically well-defined, whereas the IM is non-causal since it solves an inverse problem where the causes (actions) and effects (states) are temporally reversed.
Mathematically, inverse kinematics is an ill-posed problem in general since, in redundant robots, many actions (solutions) can lead to the desired state.
As explained in Section~\ref{sec:causal-models}, in our case, the IM is much easier to estimate since it is used for relating the pairs of consecutive states by mediating actions.

As mentioned in Introduction,
causal cognition in robots has been proposed to include a range of categories, varying in terms of complexity \cite{Hellstroem2021}. 
Here, we focus on the lower end of this spectrum and discuss low-level causality regarding two categories: sensorimotor self-learning (C1) and learning the consequences of one's own actions on objects in the environment (C2).

Our work was motivated by \cite{Lee2021}, where the authors present CREST, an approach for causal reasoning in simulation to learn the relevant state space for a robot manipulation policy.
In their approach, they conduct interventions using internal models (with simplified assumptions) that elicit the structure between the state and action spaces, enabling the construction of neural network policies with only relevant state features as input. 
They have shown on two representative manipulation tasks (block stacking and crate opening) that the policies were more robust to domain shifts, more sample-efficient to learn, and scaled to more complex settings with larger state spaces. 
CREST was presented as one approach to a broader methodology of structure-based transfer learning from simulation as a new paradigm for sim-to-real robot learning, i.e., structural sim-to-real.
The CREST achieves dimensionality reduction for reinforcement learning in the state space; the related work demonstrates that in the action space \cite{Nouri2010}.

\section{Methods}

\subsection{Data Generation}
\label{sec:data-gen}

We collected sensorimotor data in a simulated environment using the myGym toolkit \cite{Vavrecka2021}. In each step, the agent (robotic arm) executes a randomly selected action and observes a new state. Motor babbling is a natural process observed in infants during their first months. In the case of interaction with objects, the concept of intuitive physics becomes relevant. In the case of C1, the arm performs motor babbling and records its joint configuration and Cartesian effector position before and after the execution of an action.

In the case of C2, an object is added to the table in the simulated environment, and the arm has the possibility to interact with it using constrained motor babbling. During an episode, the agent observes potential changes in position, rotation and other defined features of the object, arm and environment in response to the arm's actions.

\subsection{Forward and Inverse Models}
\label{sec:causal-models}

The generated data is used for learning of the forward and inverse models. 
The FM is implemented by a feed-forward neural network that learns the mapping 
\begin{equation}
\label{eq:fm_def}
\mathrm{FM} \colon \left[ \bm{s}(t), \bm{a}(t) \right] \mapsto \bm{\hat{s}}(t+1),
\end{equation}
where the state vector features $\bm{s}(t)$ are task-dependent. In our experiments, $\bm{\theta}(t), \bm{\mathit{ef}}(t) \subset \bm{s}(t)$, where $\bm{\theta}(t)$ is the joint configuration and $\bm{\mathit{ef}}(t)$ is the effector position in Cartesian 3D space.
Here, we assume the robot perception $\bm{s}(t)$ is reliable, serving as ground truth the agent learns to estimate.\footnote{It should be acknowledged that this reliability assumption holds well in simulation but may not in real robots, when the perception may be inaccurate or even fail.}
We also assume that the action vector is represented as $\bm{a}(t) = \bm{\theta}(t+1) - \bm{\theta}(t)$, making our approach biologically plausible since the action depends on a current state.\footnote{A commonly used alternative in robotics is to represent an action as a corresponding absolute target joint vector.}

\begin{figure}[t]
  \centering
  \includeinkscape[width=0.45\textwidth]{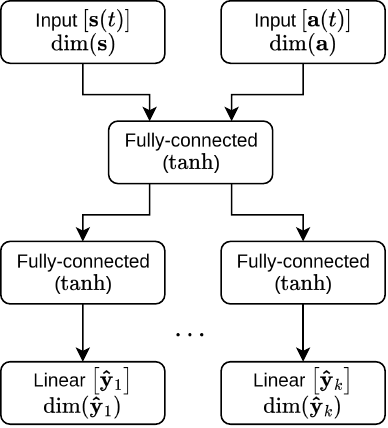}
  \caption{General forward model architecture. The linear layer denotes a fully connected layer with linear activation; hidden layers are $\tanh$-activated. $\dim(\placeholder)$ in the I/O layers designates their size.}
  \label{fig:fm_arch}
\end{figure}

Since the state features can be diverse, the architecture of the FM (Figure~\ref{fig:fm_arch})\footnote{Note that the sizes of hidden layers are task-specific based on the sizes of input and output layers. Thus, we do not list them in this general description.} contains separate output heads for different state subvectors $\bm{\hat{y}}_i \subseteq \bm{\hat{s}}(t+1)$. During the training, loss is computed separately for each subvector, and the model is optimized according to an equally weighted sum of partial losses.

\begin{figure}[h]
  \centering
  \includeinkscape[width=1\textwidth]{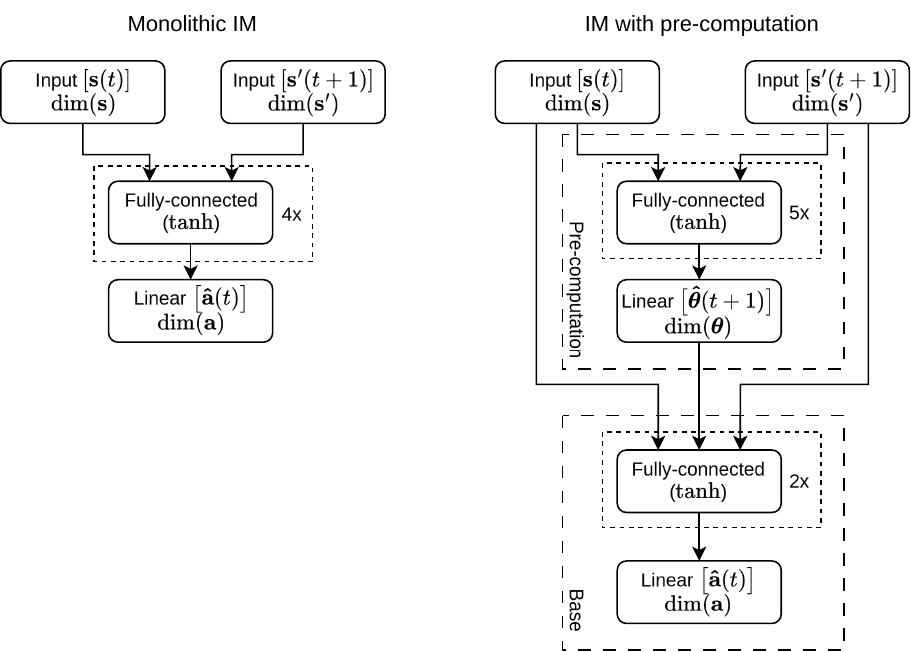}
  \caption{Architectures of monolithic IM and IM with $\bm{\theta}(t+1)$ pre-computation pre-network. The linear layer denotes a fully connected layer with linear activation, and hidden layers with ($\tanh$) are $\tanh$-activated. Multipliers next to some layers (e.g., 4x) denote the number of stacked layers of the same type. $\dim(\placeholder)$ in some layers designates their size.}
  \label{fig:im_archs}
\end{figure}

Similarly, the IM is implemented by a feed-forward neural network that learns the mapping 
\begin{equation}
\label{eq:im_def}
\mathrm{IM} \colon \left[ \bm{s}(t), \bm{s}(t+1) \right] \mapsto \bm{\hat{a}}(t).
\end{equation}
During the offline training from the generated dataset, we can assume the availability of $\bm{\theta}(t+1) \subset \bm{s}(t+1)$, but not during the inference. For this reason, we propose two approaches to the construction of inverse models.
The monolithic approach (on the left of Figure~\ref{fig:im_archs}) consists of one neural network taking $\bm{s}(t)$ and $\bm{s}'(t+1) = \bm{s}(t+1) \setminus \bm{\theta}(t+1)$ (i.e., the original state vector without $\bm{\theta}$ subvector) as input during both training and inference.

Since learning the IM mapping (Eq.~\ref{eq:im_def}) is less difficult with $\bm{\theta}(t+1)$ available, the second approach (on the right of Figure~\ref{fig:im_archs}) relies on the composition of the base model learning such mapping and the pre-network learning the mapping $\left[ \bm{s}(t), \bm{s}'(t+1) \right] \mapsto \bm{\hat{\theta}}(t+1)$ on the generated data and pre-computing the approximate value of $\bm{\theta}(t+1)$ during the inference. This output is then concatenated with the rest of the inputs and fed into the base model.

This approach aims to leverage the $\bm{\theta}(t+1)$ component during the training from the dataset but leave it free to vary unrestricted and conform to other variables during the inference. This approach may conceptually resemble the biologically relevant uncontrolled manifold (UCM) hypothesis \cite{Scholz1999}, stating that the central nervous system controls only variables pertinent to the motor task being executed and includes others in UCM to leave them uncontrolled while trying to preserve the stability of the motor action execution.

We compare both approaches in Section~\ref{sec:kinematics-exp}.

\subsection{Knowledge Extraction}
\label{sec:cm-analysis}

Trained FM can be analyzed by extracting information about the original environment and a learning session. Our primary focus is on analyzing feature importance, which allows us to highlight state features that cannot be manipulated by the agent actions and thus can be removed, hence reducing the dimensionality of the state space for the specific task and environment.
Recent related work by Lee et al. \cite{Lee2021}, which served as an inspiration, focused on determining relevant state features by conducting an intervention on one feature at a time and testing whether the same policy execution led to successful task completion or not. This way, causal dependencies were found.

On the contrary, we do not study causality by direct interaction with an environment but by using trained causal models as proxies containing this information. Using the learned FM, we can determine the relevance of state features in relation to action features by analyzing their importance.

Explaining the behaviour of trained deep neural networks is hard since these networks derive their decisions using a large number of elementary operations. Various approaches have been proposed \cite{Gilpin2018}, and one category that we focus on here is based on saliency mapping. 
Model predictions are typically based on attributions to several features (inputs). SHAP framework \cite{Lundberg2017} unifies different additive feature attribution methods.
Here, we were experimenting with Kernel SHAP and Deep SHAP. While Kernel SHAP is model-agnostic and utilizes Linear LIME \cite{Ribeiro2016}, Deep SHAP is applicable only to neural models and uses the attribution rules of the DeepLIFT method \cite{Shrikumar2017}. Both methods try to approximate the Shapley value of each input feature in relation to an output feature. Shapley value represents an input feature contribution to the output feature prediction. In the context of our work, we can use Shapley values to determine the contribution of each action variable to each state variable for a specific prediction.

In our experiments (Section~\ref{sec:physics-exp}), we use the Deep SHAP method only, as the Kernel SHAP method is significantly slower due to it making no assumptions about the analyzed model.
While the SHAP methods are local, providing an explanation for one prediction, thanks to their properties, these local explanations can be aggregated across the set of instances, providing us with global feature importance within the analyzed model, visualizable using heat maps (e.g., Figure~\ref{fig:shap_heatmap}).

Additionally, we can use partial dependence plots (PDPs) \cite{Friedman2001} to visualize the distribution of the contribution of an action feature to a state feature across the set of instances (e.g., Figure~\ref{fig:shap_pdp}). From PDPs, we can determine whether there is any relationship between the selected action and the state features and properties of this relationship. High correlation indicates a strong impact of the action feature on the state feature (however, such a relationship should not always be regarded as causal \cite{Dillon2021}).

\section{Experiments}

We applied the described methods in two experiments related to categories C1 and C2 of learning causal relations \cite{Hellstroem2021}.
The structure of both experiments is similar. The first step consists of data generation in a simulated robotic environment (Section~\ref{sec:data-gen}) provided by the myGym toolkit \cite{Vavrecka2021}.
The generated data is subsequently used for training forward and inverse models (Section~\ref{sec:causal-models}), which are further analyzed to reveal causal relations.

\subsection{Learning Kinematics}
\label{sec:kinematics-exp}

In Experiment 1, which focused on sensorimotor learning, we used KUKA LBR iiwa robotic arm with a magnetic endpoint and 7 DoF that performed motor babbling.

{\bf Environment}
The simulation\footnote{A short recording of the experiment can be seen at \url{https://youtu.be/92CcFNeFEzo}.} ran in 500,000 steps. In each step $t$, a joint configuration $\bm{\theta}(t) \in \mathbb{R}^7$ is sampled from the normal distribution with limits according to Table~\ref{tab:kuka-joints}. The magnetic endpoint is not used during this experiment.
A motor command is executed in 10 substeps before proceeding to the next simulation step, allowing a longer execution time, resulting in the actual action being more similar to the planned one. After the action execution, only the resulting configuration and Cartesian effector position $\bm{\mathit{ef}}(t) = [\mathit{ef}_x, \mathit{ef}_y, \mathit{ef}_z]$ is recorded, both composing state vector $\bm{s}(t) = \left[\bm{\theta}(t), \bm{\mathit{ef}}(t)\right]$.

\begin{table}[t]
\centering
\caption{Joint motion range of KUKA LBR iiwa arm used for Experiments 1 and 2.}
\label{tab:kuka-joints}
\begin{tabular}{@{}crrrrrrr@{}}
\toprule
& \multicolumn{1}{c}{$\theta_0$} & \multicolumn{1}{c}{$\theta_1$} & \multicolumn{1}{c}{$\theta_2$} & \multicolumn{1}{c}{$\theta_3$} & \multicolumn{1}{c}{$\theta_4$} & \multicolumn{1}{c}{$\theta_5$} & \multicolumn{1}{c}{$\theta_6$} \\ \midrule
$q_\mathit{\rm min}$ {[}rad{]} & -2.967 & -2.094 & -2.967 & -2.094 & -2.967 & -2.094 & -3.054 \\
$q_\mathit{\rm max}$ {[}rad{]} &  2.967 &  2.094 &  2.967 &  2.094 &  2.967 &  2.094 &  3.054 \\
\bottomrule
\end{tabular}
\end{table}

{\bf Models} 
The FM for this experiment uses two separate output heads, one for joint configuration prediction and the other for effector position prediction. Each head computes a separate mean squared error used as a loss. The model is trained according to the overall loss calculated as a sum of equally weighted head losses.
For training the FM, we used Adam optimizer \cite{Kingma2014} with an initial learning rate $\eta = 10^{-3}$ for 60 epochs. The model was evaluated using 5-fold cross-validation with average mean absolute error (MAE) for the effector position and joint configuration outputs of 6.9 mm and $3.4 \times 10^{-3}$ rad, respectively.

For the IM, we tried both architectural approaches proposed in Section~\ref{sec:causal-models}. All models of both approaches use mean squared error (MSE) as a loss function. The pre-computation approach uses a base IM trained first. We experimented with two variants of this model, differing in the unit's activation function at the hidden layer: hyperbolic tangent or ReLU. However, the differences in resulting performance were insignificant. 

Furthermore, we wanted to investigate whether the base IM could be input with other values for $\bm{\theta}(t+1)$ instead of estimating the true $\bm{\theta}(t+1)$. This would allow us to eliminate the pre-computation step by generating values for $\bm{\theta}(t+1)$ via a less computationally expensive process. We tested the base IM by inputting $\bm{\theta}(t + 1) = \bm{0}$ or sampling $\theta_i(t + 1)$ from the dataset. However, sampling from the dataset and inputting zero vector for $\bm{\theta}(t + 1)$ resulted in an MAE of 0.371 rad and 0.322 rad, respectively. Thus, we conclude this alternative approach is not feasible as it yields significantly worse results than the monolithic and pre-computation approaches.

Finally, we trained the feature-generating pre-network separately. After the training, the pre-network was put in front of the base model, forming the assembly used for inference. The training hyperparameters and final results of both approaches are shown in Table~\ref{tab:kuka-inv-results}.

\begin{table}[h]
\centering
\caption{Hyperparameters and resulting MAE of approaches to inverse model construction for kinematics data. Results were obtained using 5-fold cross-validation. $\eta$ and $\lambda$ denote an initial learning rate and initial weight decay, respectively, of Adam \cite{Kingma2014} and AdamW \cite{Loshchilov2017} optimizers.}
\label{tab:kuka-inv-results}
\begin{tabular}{@{}lllll@{}}
\toprule
Approach                         & Model       & Epochs & Optimizer                                  & MAE {[}rad{]}          \\ \midrule
Monolithic                       & Base        & 1,000  & AdamW($\eta = 10^{-3}$, $\lambda = 0.004$) & 0.0529                 \\ \midrule
\multirow{3}{*}{Pre-computation} & Base        & 100    & Adam($\eta = 10^{-3}$)                     & $3.06 \times 10^{-4}$  \\
                                 & Pre-network & 4,000  & AdamW($\eta = 10^{-3}$, $\lambda = 0.004$) & 0.0481                 \\
                                 & Assembly    & N/A    & N/A                                        & 0.0481                 \\ \bottomrule
\end{tabular}
\end{table}

\begin{figure}[h]
    \centering
    \includeinkscape[width=\textwidth]{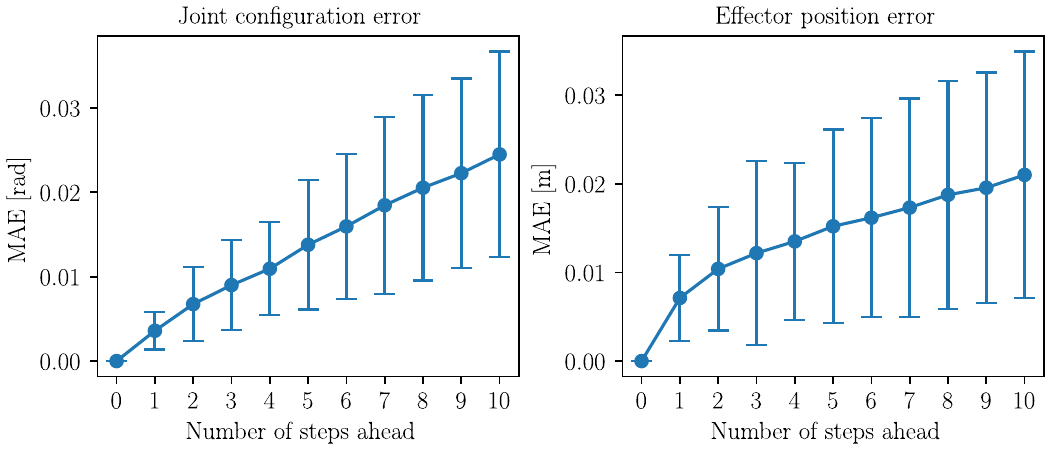}
    \caption{Average MAE and its standard deviation of joint configuration and effector position prediction by the forward model during mental simulation 10 steps ahead.}
    \label{fig:kuka_mental_sim}
\end{figure}

{\bf Mental simulation} 
We subjected the FM to the chained inference, which consisted of repeatedly querying the model on the previously generated state and random action. Provided the initial state $\bm{s}(0) \equiv \bm{\hat{s}}(0)$ and the action sequence $\tau_a = \left[\bm{a}(0), \bm{a}(1), \ldots, \bm{a}(T)\right]$ are available, we can inductively predict a successor state from the previous state as $\bm{\hat{s}}(t) = \mathrm{FM}\left[\bm{\hat{s}}(t-1), \bm{a}(t-1)\right]$. This way, we can predict the state of the environment $T$ steps ahead.

As part of Experiment 1, we evaluated the capacity of the trained FM to predict the state up to 10 steps ahead. First, we generated 6,000 random ground-truth trajectories in the simulation. Then, the first state $\bm{s}(0)$ and the action sequence of each trajectory are used in generating $\bm{\hat{s}}(t)$, with $1 \leq t \leq T$. Each predicted state $\bm{\hat{s}}(t)$ is then compared with its ground-truth counterpart $\bm{s}(t)$, and the MAE of the prediction is calculated for joint configuration and effector position subvectors separately. The evaluation results can be seen in Figure~\ref{fig:kuka_mental_sim}. The joint configuration prediction error exhibits linear growth with the number of steps ahead, while the effector position error grows in an approximate semi-logarithmic log-linear trend. The result is an improvement over our initial expectation of both errors growing exponentially.

\subsection{Simple Intuitive Physics}
\label{sec:physics-exp}

This experiment has the purpose of evaluating the proposed methods on the category 2 causal skill proposed by Hellström \cite{Hellstroem2021}, described as "Learning about how the robot affects the world".

{\bf Environment} 
The experiment\footnote{A short recording of the experiment can be seen at \url{https://youtu.be/Q_JGOD7597k}} utilizes the KUKA LBR iiwa robotic arm with a magnetic endpoint as an agent, same as in Experiment 1. The task consisted of the arm randomly switching the magnet. If it was not holding anything, the arm navigated to the magnetized cube lying on the table, picked it up, and randomly manoeuvred with it in the space for a random duration. After that, the magnet was turned off, the cube was released, and the arm babbled empty-handed for a random duration.

The goal of this experiment was to let the agent learn the simple physics of the cube as well as its kinematics. Knowledge gained by the agent in this experiment can be thus understood as a superset of knowledge from the kinematics experiment (Section~\ref{sec:kinematics-exp}). Additionally, we wanted to verify whether the model architectures specified in Section~\ref{sec:causal-models} would efficiently work with state spaces of higher dimensionality.

The data-generating simulation ran in 4,000 episodes, lasting 500 iterations each. After each iteration, we recorded the final joint configuration $\bm{\theta}(t) \in \mathbb{R}^7$, effector position and rotation as a 6D pose $\bm{\mathit{ef}}(t) = \left[ \mathit{ef}_x, \mathit{ef}_y, \mathit{ef}_z, \mathit{ef}_{\it rx}, \mathit{ef}_{\it ry} \mathit{ef}_{\it rz} \right]$, object information (its position, rotation and color) $\bm{\mathit{o}}(t) = [ o_x, o_y, o_z, o_{\it rx}, o_{\it ry}, \allowbreak o_{\it rz},  o_R, o_G, o_B ]$, and the magnetic endpoint state $\mathit{mgt}(t)$. The full state vector is composed as $\bm{s}(t) = \left[\bm{o}(t), \bm{\theta}(t), \bm{\mathit{ef}}(t), \mathit{mgt}(t)\right]$. Object colour features were added as control variables, randomized at the start of each episode, and did not change during it. 

{\bf Models}
Same as in the previous experiment, we trained an FM and a monolithic IM on the generated data.
The FM uses separate output heads for object position, object rotation, colour, joint configuration, effector position and rotation prediction. Each head computes a separate MSE, which is used as a loss. The FM was trained for 100 epochs using Adam optimizer \cite{Kingma2014}, with the initial learning rate $\eta = 10^{-3}$. For the final results of this model, see Table~\ref{tab:kuka-fwd-results}. With regard to the previous experiment, we can compare effector position and joint configuration prediction errors. While the effector position error in this experiment is slightly higher, the joint configuration error is ca. 2.5 times worse.

\begin{table}[h]
\centering
\caption{Errors of respective output heads of the forward model for intuitive physics data. Results were obtained using 5-fold cross-validation.}
\label{tab:kuka-fwd-results}
\begin{tabular}{@{}ll@{\hskip 1cm}ll@{}}
\toprule
Output head     & MAE        & Output head         & MAE                  \\ \midrule
Object position & 0.0089 m   & Effector position   & 0.008 m              \\
Object rotation & 0.0721 rad & Effector rotation   & 0.0625 rad           \\
Object color    & 0.004      & Joint configuration & 0.0084 rad           \\
                &            & Magnet state        & $1.3 \times 10^{-4}$ \\ \bottomrule
\end{tabular}
\end{table}

In this experiment, we applied only a monolithic approach to the inverse model construction as it performs similarly to the pre-computation approach but is computationally less expensive. The model is optimized according to a separate MSE for joint and magnet action prediction. The training was facilitated by AdamW optimizer \cite{Loshchilov2017} with the initial learning rate $\eta = 10^{-3}$ and initial weight decay $\lambda = 0.004$ for 1,000 epochs with the final MAE of joint action prediction 0.0077 rad and of magnet action prediction $4.56 \times 10^{-4}$.

{\bf Knowledge extraction}
The trained forward model is further analyzed using methods proposed in Section~\ref{sec:cm-analysis}. The analysis was performed using a sample of 200 observations from the generated dataset.

\begin{figure}[t]
    \centering
    \includeinkscape[width=0.9\textwidth]{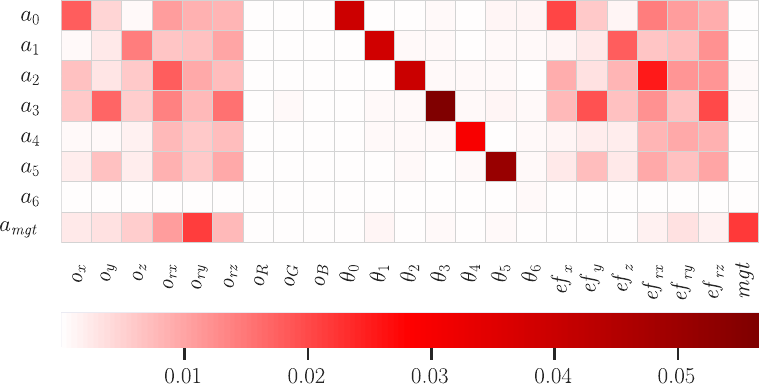}
    \caption{Contribution heat map generated by Deep SHAP method on the forward model showing the magnitude of the contribution of specific actions to output features.}
    \label{fig:shap_heatmap}
\end{figure}

The resulting global contribution heat map generated for this experiment is shown in Figure~\ref{fig:shap_heatmap}. The y-axis denotes the action of each joint and a magnetic endpoint of the arm. The x-axis contains defined environment state features. The colour of each cell corresponds to the magnitude of contributions of action features to the state features averaged across the selected sample.

The figure shows, for instance, that joint 6 is not used during the simulation as its action ($a_6$) does not correlate with the change of any state variable and, most importantly, with the change of the state of the joint itself. In addition, the colour of the object is irrelevant in this case, as no action can affect it and thus could be removed (or ignored) from the state space. On the other hand, all action features, except $a_6$, affect most object features. This low-level knowledge can be useful for causal analysis at higher levels.

\begin{figure}[h]
    \centering
    \includeinkscape[width=0.49\textwidth]{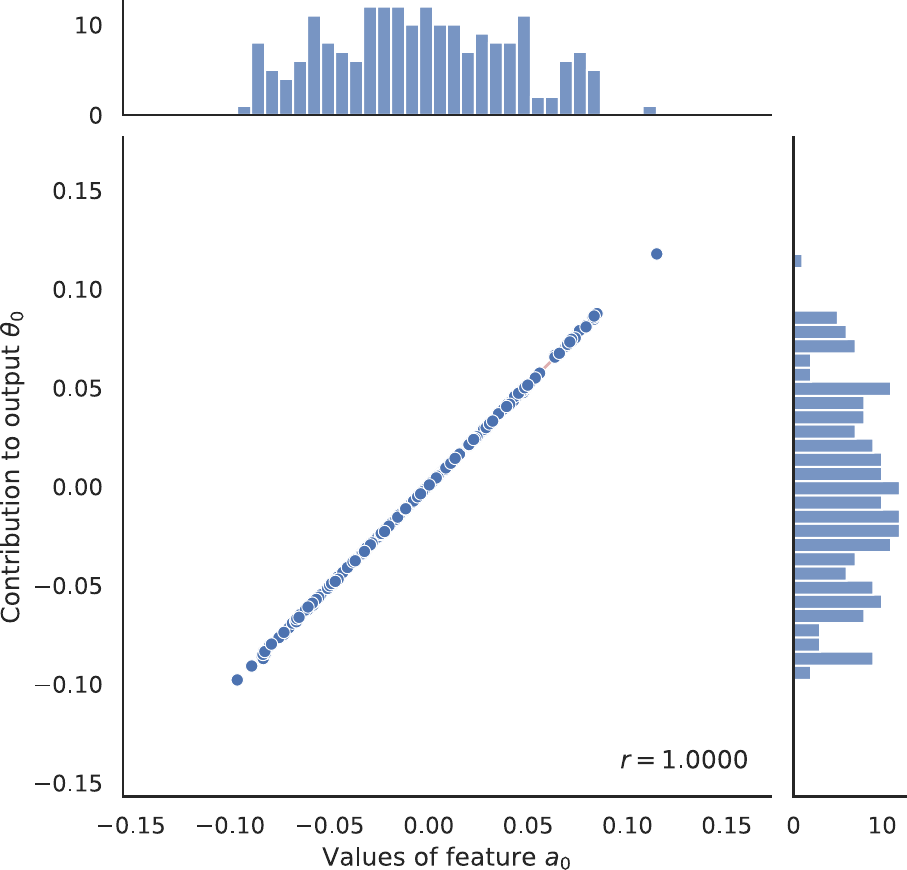} 
    \includeinkscape[width=0.49\textwidth]{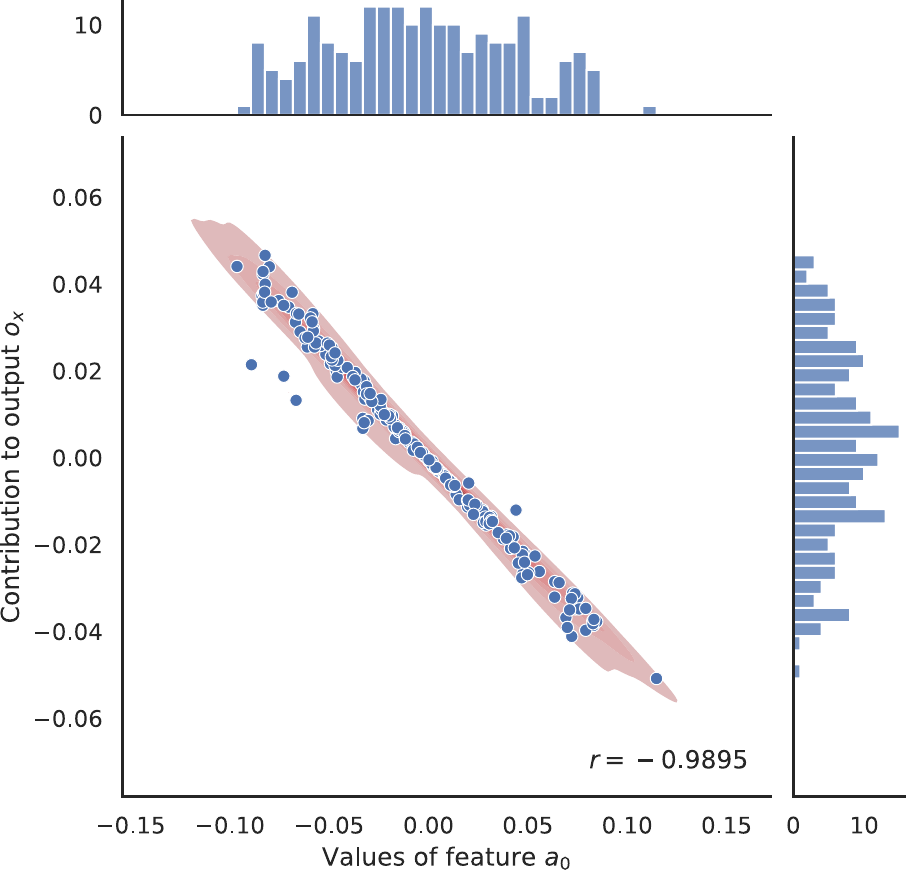}
    \includeinkscape[width=0.49\textwidth]{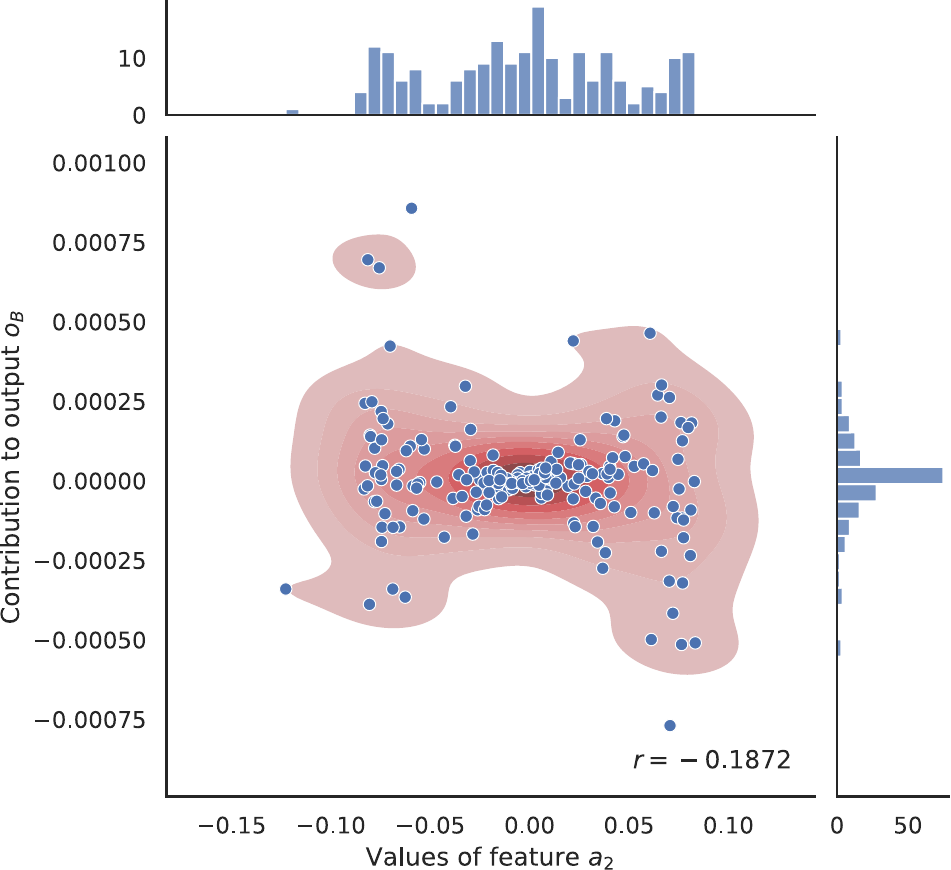}
    \includeinkscape[width=0.49\textwidth]{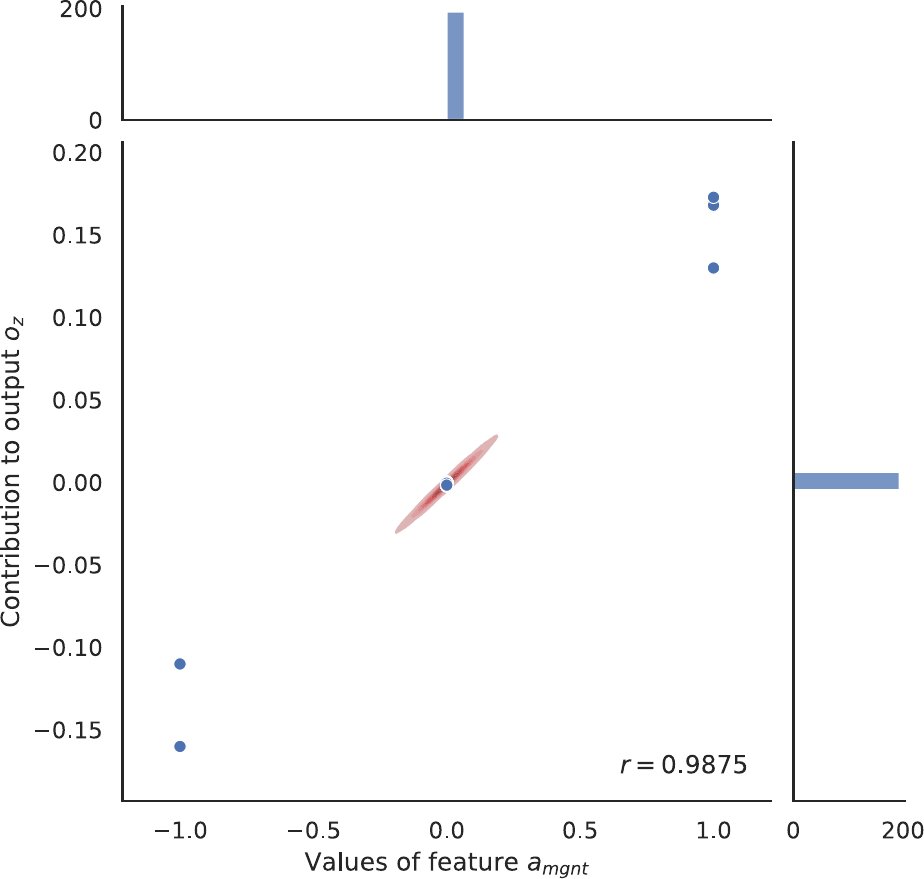}
    \caption{A sample of partial dependence plots generated by Deep SHAP method applied to the forward model showing correlation between a value of a specific action component and its contribution to an output variable.}
    \label{fig:shap_pdp}
\end{figure}

Feature importance and dependencies can also be studied using feature contribution distributions and partial dependence plots. Figure~\ref{fig:shap_pdp} shows a sample of PDPs generated from feature contribution data output by Deep SHAP method \cite{Lundberg2017}. Averaged absolute contribution across these distributions corresponds to the values in Figure~\ref{fig:shap_heatmap}. 

From the presented sample of PDPs, we can observe that action feature $a_0$ (movement of joint 0) has a substantial impact on state features $\theta_0$ (state of joint 0) and $o_x$ (object position on the x-axis). Moreover, action feature $a_2$ does not significantly correlate with the contribution to state feature $o_B$ (blue color component of the object) with contribution centered around 0.0 indicating $a_2$ does not profoundly impact $o_B$. Last, the action of the magnetic endpoint $a_{\mathit{mgt}}$ is prevalently null, as the state of the magnet does not often change between iterations. However, when it does, it significantly contributes to $o_z$ (object position on the z-axis) since turning the magnet on ($a_\mathit{mgt} = 1$) or off ($a_\mathit{mgt} = -1$) in this experiment is followed by lifting the object in the air or dropping it on the table.

\section{Conclusion}

In this paper, we explored causal relations by learning the forward and inverse models on synthetic data generated in simulation. We confirmed that the forward model constructed using the proposed approach could be successfully used for mental simulation, possibly helping with action planning by predicting future states based on causality observed in the data. Additionally, we explored approaches to inverse model construction, allowing the model to predict the action needed for a transition between subsequent states.

Moreover, we demonstrated the capability of extracting knowledge about the environment's behaviour from the trained forward model using explainability methods. We proposed that information obtained this way can be used to determine relevant state features, serving as a basis for dimensional reduction.
Our method can be applied to scenarios that are much more complex and much harder for humans to understand, and thus, it can be an important tool for extracting causal knowledge.
For future work, we plan to investigate the task of action planning as an imitation learning assisted by the proposed models.

\subsubsection{Acknowledgements.} The authors would like to thank Michal Vavrečka for his advice, feedback and technical support. This research was supported by the Horizon Europe project TERAIS, GA No. 101079338 and in part by the Slovak Grant Agency for Science (VEGA), project 1/0373/23.

\bibliographystyle{splncs04}
\bibliography{references}

\end{document}